\pdfoutput=1

\documentclass[11pt]{article}

\usepackage[]{ACL2023}

\usepackage{times}
\usepackage{latexsym}
\usepackage{url}            
\usepackage{booktabs}       
\usepackage{amsfonts}       
\usepackage{nicefrac}       
\usepackage{microtype}      
\usepackage{xcolor}         
\usepackage{latexsym}
\usepackage{amssymb}
\usepackage{amsmath}
\usepackage{booktabs}
\usepackage{enumerate}
\usepackage{graphicx}
\usepackage{subfigure}
\usepackage{xspace}
\usepackage{float}
\usepackage{bbm}
\usepackage{bm}
\usepackage{multirow}
\usepackage{booktabs}
\usepackage{color}
\usepackage{framed}
\usepackage{stfloats}
\usepackage{iitem}
\usepackage{amssymb}
\usepackage{pifont}
\usepackage{arydshln}
\usepackage{enumitem}
\usepackage{wrapfig}
\usepackage{algorithm}
\usepackage{algpseudocode}
\usepackage{makecell}

\usepackage[T1]{fontenc}

\usepackage[utf8]{inputenc}

\usepackage[most]{tcolorbox}
\usepackage{tabularx}

\usepackage{hyperref}

\usepackage{microtype}

\usepackage{inconsolata}
\usepackage{multirow}
\usepackage[T1]{fontenc}
\usepackage{colortbl}
\definecolor{maroon}{cmyk}{0,0.87,0.68,0.32}

\newcommand{\method}{APT\xspace}

\title{APT: Improving Specialist LLM Performance with Weakness Case \underline{A}cquisition and  Iterative \underline{P}reference \underline{T}raining}

\author{
Jun Rao$^{1}$,
 Zepeng Lin$^{1}$,
 Xuebo Liu$^{1}$\thanks{~~Corresponding Author},
 Xiaopeng Ke$^{1}$,
 Lian Lian$^{2}$,
 Dong Jin$^{2}$,
 \\ \bf{Shengjun Cheng}$^{2}$,
    \bf{Jun Yu}$^{3}$, and
    \bf{Min Zhang}$^{1}$\\
    \textsuperscript{\rm1}Institute of Computing and Intelligence, Harbin Institute of Technology, Shenzhen, China \\
     \textsuperscript{\rm2}Huawei Cloud Computing Technologies Co., Ltd.~~~\\
         \textsuperscript{\rm3}School of Intelligence Science and Engineering, Harbin Institute of Technology, Shenzhen, China~~~
    \\
    \texttt{\{rao7jun,zepenglin11,xiaopk7\}@gmail.com, \{liuxuebo,zhangmin2021\}@hit.edu.cn}\\
    \texttt{\{lianlian3,jindong2,chengshengjun\}@huawei.com}
    }

\begin{document}
\maketitle
\begin{abstract} 
Large Language Models (LLMs) often require domain-specific fine-tuning to address targeted tasks, which risks degrading their general capabilities. Maintaining a balance between domain-specific enhancements and general model utility is a key challenge.
This paper proposes a novel approach named APT (Weakness Case \textbf{A}cquisition and Iterative \textbf{P}reference \textbf{T}raining) to enhance domain-specific performance with self-generated dis-preferred weakness data (bad cases and similar cases).
APT uniquely focuses on training the model using only those samples where errors occur, alongside a small, similar set of samples retrieved for this purpose. This targeted training minimizes interference with the model’s existing knowledge base, effectively retaining generic capabilities.
Experimental results on the LLama-2 and Mistral-V0.3 models across various benchmarks demonstrate that APT ensures no reduction in generic capacity and achieves superior performance on downstream tasks compared to various existing methods. This validates our method as an effective strategy for enhancing domain-specific capabilities without sacrificing the model's broader applicability. 
\end{abstract}

\section{Introduction} 
\begin{figure}[t!]
    \centering
\includegraphics[width=1\linewidth]{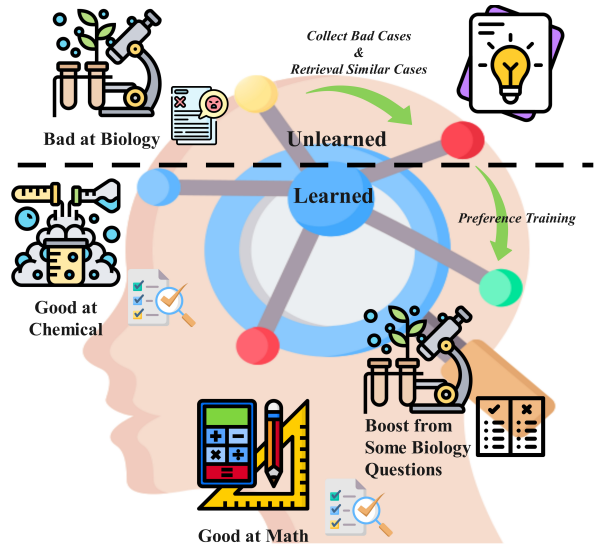}
    \caption{Similar to how people improve in weaker areas by focusing on their mistakes and practicing similar problems, our method identifies model deficiencies and uses targeted training to enhance performance.}
    \label{example}
\end{figure}


Large Language Models (LLMs) can be applied to multiple domains, such as coding~\cite{deepspeed}, mathematics~\cite{meta-math}, general knowledge question answering~\cite{dubois2024alpacafarm,pesf-kd}. 
However, substantial differences exist among these domains and tasks, and real-world applications often suffer from limited domain-specific data due to industry barriers. 
Consequently, enhancing domain models with a small amount of domain-specific data combined with general data~\cite{shi2023specialist} has emerged as a prominent research focus.
Recent work has emphasized self-training to improve task-specific capabilities, particularly for coding and mathematical reasoning tasks~\cite{gsm,step-dpo,glm_math,meta-math}. 
These studies scale the original datasets through task-specific data generation strategies, resulting in enhanced domain models. 
Other approaches aim to improve models' capabilities in specific domains without compromising general task performance, using techniques such as data mixing~\cite{dong2023abilities} and self-distillation~\cite{yang2024self}.
Existing methods typically focus on individual tasks or enhance models by constructing more instruction fine-tuned data. 
The former approach suffers from limited generalization due to its focus on specific tasks, while the latter primarily leverages information from positive samples and lacks feedback from negative samples.

{In this paper, we propose the APT method, which utilizes a small number of self-generated negative samples for iterative preference training.}
The dis-preferred data generation process does not require complex task-specific data construction. 
Similar to how humans learn from similar problems shown in Figure \ref{example}, our method has the model generate responses to tasks and add incorrect responses (``bad cases'') back into this iteration training data. Additionally, a retrieval module identifies similar samples (``similar cases'') to add more relevant training data. {These two types of data are then used for training, which enables the model to learn from its errors and generalize across related tasks, effectively improving its performance on specific tasks.}
The model training component involves constructing preference data, where model-generated responses are non-preference data and original correct answers are preference data. 
Through {iterative} preference learning, the model optimizes its output by recognizing and correcting errors while learning from similar samples. This approach strengthens the model's ability to handle challenging cases, ultimately improving performance.

Through extensive experimental validation, we show that APT is effective in identifying and optimizing model flaws across several domains, such as mathematical reasoning, coding, and instruction-following. 
The results indicate improvements of up to 5.9\% in Llama2-7b and 6.0\% in Mistral-7b, outperforming existing methods such as \citet{dong2023abilities}. 
Additionally, ablation studies show that acquiring bad case data, retrieving similar cases, and optimizing preference data training objectives and iterations all contribute to enhancing model performance. 
Final scaling experiments demonstrate the method's potential for future improvements on stronger models.
Our contributions are as follows:
\begin{itemize}
    \item We present APT, the first preference optimization framework that combines the model's self-generated data of bad cases and retrieved relevant data to further enhance the model through iterative preference training.

    \item APT shows the ability to enhance special tasks while ensuring that general-purpose tasks (seven large-scale datasets) are not degraded and has been validated on a number of specific domains, including mathematical reasoning, coding, and instruction following. 

    \item We find that LLMs improve through iterative preference training by learning from retrieval of similar mistakes, much like humans, and even in much stronger domain models.

    \end{itemize}

\section{Related Work}
\subsection{Alignment Finetuning}
Alignment learning trains models to align outputs with human preferences or goals, often using techniques like reward shaping or reinforcement learning to optimize performance and minimize unexpected behaviors. 
Some methods use instruction fine-tuning~\cite{xu2023baize,alpaca-gpt4,cheng2024autodetect,commonIT} to facilitate learning with high-quality examples. Others leverage contrastive data to enhance model consistency with human intentions by increasing positive responses and reducing negative ones.
\citet{dpo} propose Direct Preference Optimization (DPO) to efficiently train large models for knowledge alignment using preference rankings instead of reward models. 
DPO optimizes classification loss from preference data, making implementing it simpler than reinforcement learning from human feedback (RLHF). 
Subsequent studies have used DPO for multi-domain alignment. 
\citet{gpo} introduces Group Preference Optimization to improve alignment training further.

\subsection{Self-Training}
In the past, most training data sources consisted of existing manually curated supervised corpora. 
In contrast, as large models evolve, synthetic data demonstrates significant potential for application. 
Recent efforts leverage the generative power of GPT-4~\cite{gpt4} to synthesize various instruction fine-tuning and alignment datasets~\cite{alpaca-gpt4,DatabricksBlog2023DollyV2,xu2023baize,ImprovingSimultaneous_DDL+23,NewTermBenchmarking_DJL+24}. 
Earlier work, such as Self-Critiquing~\cite{self_critiquing}, used a model's own output to blend with the original answer distribution, enhancing model performance. 
Many recent studies~\cite{deepseekv2,qwen}, including Nemotron~\cite{nemotron}, extensively use synthetic data and have established automated synthetic data generation processes. 
\citet{instruction_evolving} employs large models to evolve the instruction fine-tuning dataset, enhancing multitasking capabilities. 
\citet{error_evolving} leveraged model output errors to evolve the model and improve its performance. \citet{chen2024self} use LLMs to generate training data from previous iterations, refining policies by distinguishing between self-generated responses and those derived from human annotations. 
Our work employs a self-play mechanism to address model weaknesses. It identifies poorly processed samples and retrieves similar samples to improve the processing of these cases.
\begin{figure*}[t]
    \centering
    \includegraphics[width=\linewidth]{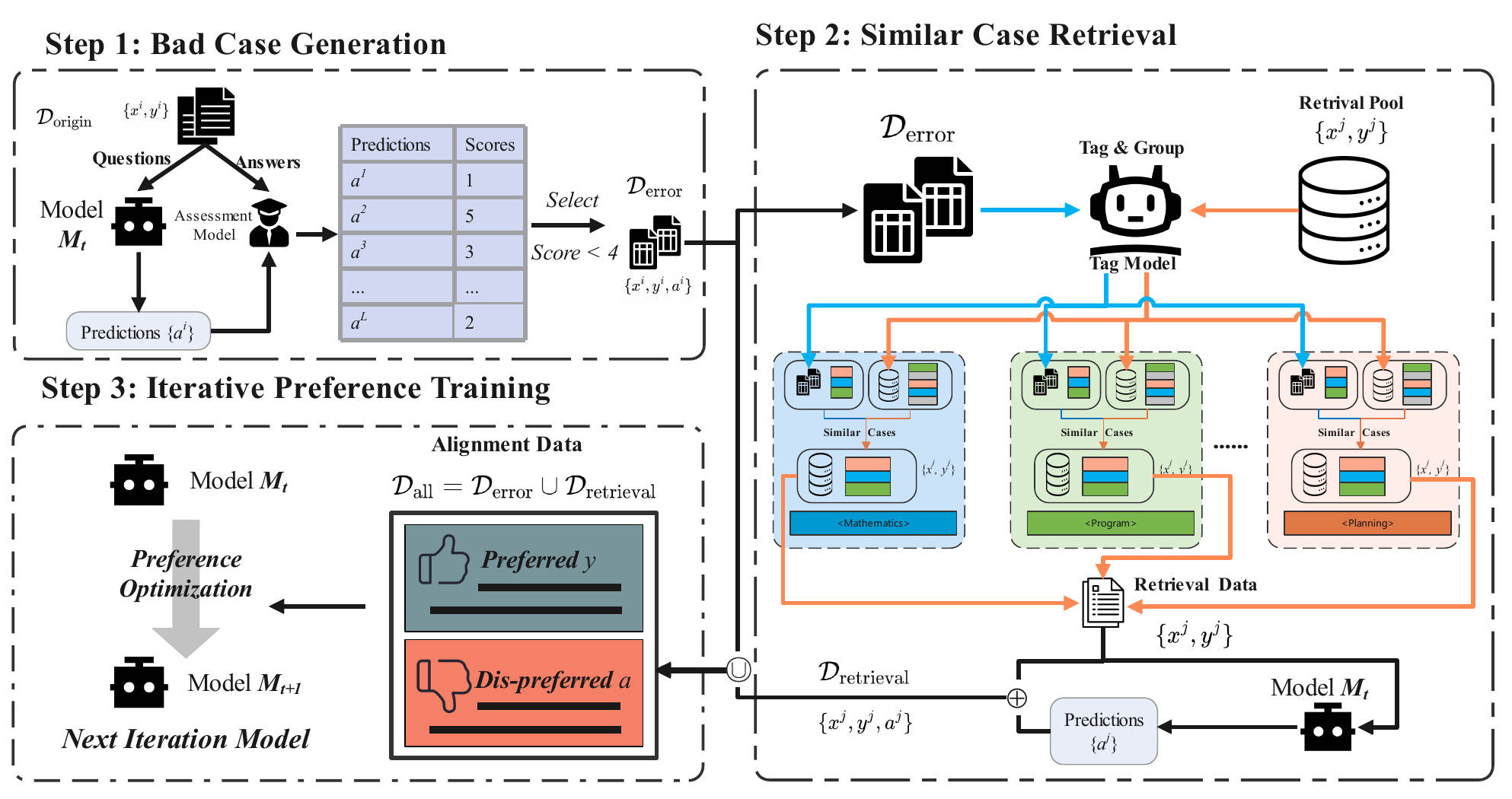}
    \caption{An overview of our approach APT. The main components of our method include: 1) Bad Case Generation: {We classify samples with low scores from the assessment model as ``Bad Cases'', which the model struggles to learn effectively. The predictions are then combined with the ground truth to create an Error Preference dataset $D_{\mathrm{error}}$ }. Appendix~\ref{Judge_prompt} presents the assesssment prompt; 2) Similar Case Retrieval: {We use retrieval to identify examples related to $D_{\mathrm{error}}$ from the retrieval pool, thus forming the dataset $D_{\mathrm{retrieval}}$ }; and 3) Iterative Preference Optimization: { $D_{\mathrm{error}}$ and $D_{\mathrm{retrieval}}$ are the final preference datasets. Appendix~\ref{sec:apt_algorithm} presents the algorithm.}}
    \label{overview}
\end{figure*}
\section{Method}

\subsection{{Overview}} \label{sec:overview}
As illustrated in Figure \ref{overview}, we begin by defining the components involved in this iterative process:
\begin{itemize}
     \item \textbf{Initial Model}: Let the existing model be denoted as $M_t=f_{\theta}$, where $\theta$ represents the model parameters. This model serves as the starting point for subsequent iterations. {Additionally, let $f_{\theta_0}$ be the assessment model (Prometheus 2-7B)~\cite{kim2024prometheus}, which is used to evaluate the performance of $M_t$ throughout the iterative process.}  
     \item 
     \textbf{Initial Dataset}: We define the task-specific supervised fine-tuning (SFT) dataset as $\mathcal{D}_{\mathrm{origin}} = {(x^i, y^i)}_{i=1}^L$, where each pair $(x^i,y^i)$ consists an input-output example used for model generation and identifying bad cases. \item 
     \textbf{Retrieval Pool}: We define the retrieval pool we constructed as $\mathcal{D}_{\mathrm{pool}} = {(x^j, y^j)}_{j=1}^U$. \item 
    \textbf{Preference Dataset}: {The error dataset $\mathcal{D}_{\mathrm{error}} = \{(x^i, y^i, a^i)\}_{i=1}^N$ is constructed by treating the model's prediction $a^i$ as dis-preference and the corresponding ground truth $y^i$ from $\mathcal{D}_{\mathrm{origin}}$ as preference. 
    The samples in $\mathcal{D}_{\mathrm{error}}$ are specifically selected from those cases where the assessment model $f_{\theta_0}$ assigns a low score.} The dataset $\mathcal{D}_{\mathrm{retrieval}} = \{(x^j, y^j, a^j)\}_{j=1}^M$ is formed by organizing the similar data retrieved during the retrieval phase similar to $\mathcal{D}_{\mathrm{error}}$. 
    {Finally, we obtain the preference dataset $\mathcal{D}_{\mathrm{all}} = \mathcal{D}_{\mathrm{error}} \cup \mathcal{D}_{\mathrm{retrieval}}$.} 
    
\end{itemize}



\subsection{Bad Case Generation
}

{Similar to the human error correction process, LLMs can improve their weaknesses by actively learning from challenging samples and enhancing their most vulnerable areas.} Based on this idea, we use initial model $M_t$ to self-predict the questions $\{x^i\}_{i=1}^L$ from the SFT training dataset $\mathcal{D}_{\mathrm{origin}}$ and obtain the corresponding predictions $\{a^i\}_{i=1}^L$. 
{These predictions, along with their corresponding ground truth answers $\{y^i\}_{i=1}^L$ and the original questions $\{x^i\}_{i=1}^L$, are evaluated using the Assessment Model $f_{\theta_0}$. For each instance, the score $s^i = f_{\theta_0}(x^i,y^i,a^i)$ quantifies how well the response of the model aligns with human evaluations. 
By selecting data instance $(x^i,y^i,a^i)$ with low score $s^i$, we identify cases where the model struggles and fails to handle the data effectively.} 



\subsection{Similar Case Retrieval}
Bad cases are relatively rare due to the strong capabilities of specialist models. 
This scarcity may lead to overfitting during model training. 
Existing approaches to address this include constructing similar question-answer pairs by model-generated rewrites~\cite{step-dpo}, regenerating similar questions~\cite{cheng2024autodetect}, and employing back-translation~\cite{meta-math}. 
However, these methods require further design and filtering for specific tasks and lack generalizability. 
In this work, we adopt a more direct strategy: retrieving data related to the current bad case using the existing high-quality instruction data as the retrieval source.

\paragraph{{Retrieval Pool Construction}}
The structure of data retrieval pool $\mathcal{D}_{\mathrm{pool}}$ is as follows: we follow the configuration of \citet{wang2023far} to collect a comprehensive and representative sample of datasets across various styles. These datasets include: (1) existing NLP resources, such as SuperNI~\cite{supernli} and Flan V2~\cite{flan_v2}; (2) datasets created from scratch by humans specifically for instruction tuning, such as Lima~\cite{zhou2023lima}; (3) datasets generated by proprietary models, including Self-Instruct~\cite{wang2022self}, Unnatural Instructions~\cite{honovich-etal-2023-unnatural}, GPT4-Alpaca~\cite{alpaca-gpt4}, Baize~\cite{xu2023baize}, OpenOrca~\cite{OpenOrca}, WizardLM~\cite{xu2023wizardlm}, and Meta-Math~\cite{meta-math}; and (4) datasets designed for specialized skills, such as CoT~\cite{cot} for chain-of-thought reasoning.

\paragraph{Tag-Based Retrieval}
{
We propose the \textsc{Tag-Based Similarity} method, which leverages a tagging model~\cite{lu2023instag} to enhance retrieval precision by categorizing data into finer-grained classes and performing tag-specific similarity searches. Tag statistics are provided in Appendix~\ref{statis_tag_info}, and the detailed procedure is as follows:}
{Given an error case preference dataset $\mathcal{D}_{\mathrm{error}}$ and a retrieval pool $\mathcal{D}_{\mathrm{pool}}$, we apply the tagging model to the input $x$ of each data point. Specifically, the tagging model assigns one or more tags $\mathcal{T}(x)$ to each input $x$, representing its semantic categories such as ``Mathematics'' or ``Program''. For each tag $t \in \mathcal{T}$, we extract subsets:}

{\small
\setlength{\abovedisplayskip}{4pt}  
\setlength{\belowdisplayskip}{4pt}  
\begin{gather}
\mathcal{D}_{\mathrm{error}}^{t} = \{(x^i, y^i, a^i) \in \mathcal{D}_{\mathrm{error}} \ : t \in \mathcal{T}(x^i,y^i) \} ,\label{eq:error} \\
\mathcal{D}_{\mathrm{pool}}^{t} = \{(x^j,y^j) \in \mathcal{D}_{\mathrm{pool}} \ : t \in \mathcal{T}(x^j,y^j) \}, \label{eq:pool}
\end{gather}
}

where $t$ represents a semantic category assigned by the tagging model. Consequently, this process produces multiple subsets, $(\mathcal{D}_{\mathrm{error}}^{t}, \mathcal{D}_{\mathrm{pool}}^{t})$, each associated with a specific tag $t$.
{For each pair of subsets $(\mathcal{D}_{\mathrm{error}}^{t}, \mathcal{D}_{\mathrm{pool}}^{t})$, we use the Sentence Transformer\footnote{Here, we use the all-MiniLM-L6-v2 model to generate embedding vectors.}~\cite{reimers-gurevych-2019-sentence} to generate embedding vectors for the input $x$ of each data point, resulting in ${\mathbf{e}(x^i)}_{i=1}^N$ and ${\mathbf{e}(x^j)}_{j=1}^U$, respectively. Next, for the subset $\mathcal{D}_{\mathrm{error}}^{t}$, we compute the average embedding vector:}
\begin{equation}
    \mathbf{e}_{\mathrm{avg}}^{t} = \frac{1}{|\mathcal{D}_{\mathrm{error}}^{t}|} \sum_{(x^i, y^i, a^i) \in \mathcal{D}_{\mathrm{error}}^{t}} \mathbf{e}(x^i).
\end{equation}

{Then, the cosine similarity between $\mathbf{e}_{\mathrm{avg}}^{t}$ and the embedding vectors of the samples in $\mathcal{D}_{\mathrm{pool}}^{t}$ is computed as follows:}
\begin{equation}
    S(\mathbf{e}_{\mathrm{avg}}^{t}, x^j) = \frac{\mathbf{e}_{\mathrm{avg}}^{t} \cdot \mathbf{e}(x^j)}{\|\mathbf{e}_{\mathrm{avg}}^{t}\|\|\mathbf{e}(x^j)\|}
\end{equation}

{Based on the computed similarity scores, we select the most similar samples from $\mathcal{D}_{\mathrm{pool}}^{t}$ such that their number matches the size of $\mathcal{D}_{\mathrm{error}}^{t}$, forming a retrieved subset $\mathcal{D}^{t}_{\mathrm{retrieval}}$. Finally, we aggregate the retrieved subsets across all tags to construct the complete retrieval set:}
\begin{equation}
    \mathcal{D}_{\mathrm{retrieval}} = \bigcup_{t \in \mathcal{T}} \mathcal{D}^{t}_{\mathrm{retrieval}}
\end{equation}

\begin{table*}[t]
  \centering
  \scalebox{0.75}{
    \begin{tabular}{llllllcc}
    \toprule
    \multirow{2}{*}{\textbf{Model}} &\multirow{2}{*}{\textbf{Method}} & \multicolumn{1}{p{4.225em}}{\textbf{Domain}\newline{}\textbf{Dataset}} & \multicolumn{1}{p{4.225em}}{\textbf{Math} \newline{}\textbf{Reasoning}} & \multirow{2}{*}{\textbf{Coding}} & \multicolumn{1}{p{4.225em}}{\textbf{Instruction} \newline{}\textbf{Following}} & \multicolumn{1}{p{4.225em}}{{\textbf{General} \newline{}\textbf{Capability}}} & \multirow{2}{*}{\textbf{AVG}} \\
    \midrule
        \multirow{14}{*}{\textbf{LLama2-7B}} &\multicolumn{1}{l}{\textbf{Base}} & - & 15.5 & 19.2 & 7.5 & 56.9 & 24.8 \\
    \cmidrule(l){2-8}
    &\multirow{3}[2]{*}{\textbf{SFT}} 
        & GSM        & 32.1 & 18.5 & 34.3 & 58.5 & 35.9 \\
        && CodeAlpaca & 17.7 & 23.6 & 29.2 & 58.7 & 32.3 \\
        && Dolly      & 19.3 & 20.9 & 16.9 & 59.1 & 29.1 \\
    \cmidrule(l){2-8}
    &\textbf{Mixed Training} & - & 34.1 & 25.0 & 19.1 & 59.1 & 34.3 \\
    \cmidrule(l){2-8}
    &\multirow{3}[2]{*}{\textbf{\quad+Continued SFT}} 
        & GSM        & 32.6 & 27.4 & 19.5 & 59.0 & 34.6 \\
       & & CodeAlpaca & 31.9 & 28.1 & 21.1 & 58.9 & 35.0 \\
        && Dolly      & 33.2 & 28.1 & 18.0 & 59.1 & 34.6 \\
    \cmidrule(l){2-8}
    &\multirow{3}[2]{*}{\textbf{\quad+DMT}\small{~\cite{dong2023abilities}}} 
        & GSM        & 34.0 & 26.9 & 18.5 & 59.0 & 34.6 \\
        && CodeAlpaca & 33.4 & 26.4 & 18.3 & 58.5 & 34.2 \\
        && Dolly      & 33.7 & 26.0 & 18.0 & 58.4 & 34.0 \\
    \cmidrule(l){2-8}
    &\multirow{3}[2]{*}{\textbf{\quad+Ours}} 
        & GSM        & \textbf{39.2} (+5.1)  & 26.1  & 24.2  & \textbf{59.4}  & \textbf{37.2} (+2.9)  \\
        && CodeAlpaca  & 34.8  & \textbf{28.4} (+3.4)  & 24.3  & \textbf{59.3}  & \textbf{36.7} (+2.4)  \\
        && Dolly      & 34.7  & 27.4  & \textbf{25.0} (+5.9)  & \textbf{59.3}  & \textbf{36.6} (+2.3)  \\
    \midrule     \midrule
    \multirow{14}{*}{\textbf{Mistral-7B-V0.3}}&\textbf{Base} & - & 40.6 & 33.1 & 16.4 & 64.5 & 38.7 \\
    \cmidrule(l){2-8}
    &\multirow{3}[2]{*}{\textbf{SFT}} 
        & GSM        & 56.9 & 41.1 & 33.8 & \textbf{66.5} & 49.6 \\
        && CodeAlpaca & 46.4 & 43.8 & 30.3 & 66.1 & 46.7 \\
        && Dolly      & 43.8 & 41.0 & 26.0 & \textbf{66.8} & 44.4 \\
    \cmidrule(l){2-8}
    &\textbf{Mixed Training} & - & 58.2 & 43.7 & 30.0 & 66.4 & 49.6 \\
    \cmidrule(l){2-8}
    &\multirow{3}[2]{*}{\textbf{\quad+Continued SFT}} 
        & GSM        & 58.9 & 43.6 & 29.4 & 66.3 & 49.6 \\
        && CodeAlpaca & 58.9 & 44.6 & 31.2 & 66.3 & 50.3 \\
        && Dolly      & 58.7 & 44.1 & 27.2 & 66.3 & 49.1 \\
    \cmidrule(l){2-8}
    &\multirow{3}[2]{*}{\textbf{\quad+DMT}\small{~\cite{dong2023abilities}}} 
        & GSM        & 59.3 & 44.5 & 30.1 & 66.1 & 50.0 \\
        && CodeAlpaca & 59.3 & 42.9 & 29.0 & 65.9 & 49.3 \\
        && Dolly      & 59.1 & 43.6 & 29.1 & 66.0 & 49.5 \\
    \cmidrule(l){2-8}
    &\multirow{3}[2]{*}{\textbf{\quad+Ours}} 
        & GSM         & \textbf{61.8} (+3.6)  & 45.9  & 30.6  & \textbf{66.5}  & \textbf{51.2} (+1.6)  \\
        && CodeAlpaca & 60.2  & \textbf{49.7} (+6.0)  & 31.2  & \textbf{66.7}  & \textbf{52.0} (+2.4)  \\
        && Dolly      & 59.6  & 47.6  & \textbf{35.0} (+5.0)  & \textbf{66.8}  & \textbf{52.3} (+2.7)  \\

    \bottomrule
    \end{tabular}}
    \caption{{Main results on multiple LLMs across domain and general tasks. Continued SFT, DMT, and Ours are built upon the Mixed Training model, which is undergoing further training to improve domain-specific performance. Our approach achieves the best results across all domains while maintaining excellent performance overall.}}
  \label{main}
\end{table*}

\subsection{Iterative Preference Training} 
To enhance the alignment of LLMs with human preferences, fine-tuning loss is commonly employed the following:
\begin{equation}
    {L}_\mathrm{SFT}(\boldsymbol{\theta}) = -\sum_{i = 1}^{N}y_{w_i}\log(\hat{y}_{w_i}|x)
\end{equation}


Training reward functions is challenging in practice, but DPO~\cite{dpo} simplifies this process using a predefined preference dataset, which we use $\mathcal{D}_{\text{all}}$ here.
The objective optimization function for this process is as follows:

\vspace{-10pt} 
\begin{small}
\begin{equation}
\begin{aligned}
&L_{\mathrm{DPO}}\left(\boldsymbol{\theta}, \boldsymbol{\theta}_{\mathrm{ref}}\right)=\mathbb{E}_{\left({x}, {y}, {a}\right) \sim \mathcal{D}_{\text{all }}}\\
&\left[\ell\left(\lambda \log \frac{p_{\boldsymbol{\theta}}\left({y} \mid {x}\right)}{p_{\boldsymbol{\theta}_{\text {ref }}}\left({y} \mid {x}\right)}-\lambda \log \frac{p_{\boldsymbol{\theta}}\left({a} \mid {x}\right)}{p_{\boldsymbol{\theta}_{\text {ref }}}\left({a} \mid {x}\right)}\right)
\right]
\end{aligned}
\end{equation}
\end{small}


Here, ${x} \sim D$ is sampled from a given distribution $D$, and the KL regularization term prevents excessive deviation of the new model $p_\theta$ from the reference model $p_{ref}$, with the regularization parameter $\lambda > 0$.
However, DPO suffers from a decrease in the probability of choosing, and some work has been done to improve this by using different constraints~\cite{hong2024orpo,pal2024smaug}. 
We directly use SFT loss as a constraint, and the final optimization function used in our experiments is shown below:
\begin{equation}\label{we_loss}
    {L}(\boldsymbol{\theta}) = {L}_\mathrm{DPO}(\boldsymbol{\theta},\boldsymbol{\theta}_{ref}) + \alpha {L}_\mathrm{SFT}(\boldsymbol{\theta}) 
\end{equation}
After training, we obtain an initial model for the next iteration to continue the previous data acquisition and training process.

\subsection{Efficiency Discussion}
Once generated, TAG annotations can be reused indefinitely, eliminating the need for repeated labeling. Despite processing a large volume of data, including content from the retrieval pool, the TAG model remains lightweight, with minimal token consumption, ensuring low computational overhead. In fact, the TAG process only requires an additional 5\% of the total time, which is negligible in the overall workflow. Furthermore, when a new task arrives, only TAG processing and retrieval are necessary, incurring minimal time costs, which makes our approach both efficient and sustainable.
\section{Experiment}\label{cr}
\subsection{Experimental Setup}
\paragraph{Training Datasets}
\textbf{Alpaca}~\cite{alpaca,alpaca-gpt4} is generated by the Self-Instruct framework and consists of 52K triplets, which is a common use of general domain data.
We use CodeAlpaca, GSM, and Dolly for the special domain data.
\textbf{CodeAlpaca}~\cite{codealpaca} is an instruction fine-tuning training data about the code, with about 20K samples.
\textbf{GSM8K} dataset~\cite{gsm}, which consists of 9K high-quality arithmetic word problems designed for the grade school level.
\textbf{Dolly-v2}~\cite{DatabricksBlog2023DollyV2} is the open-sourced model trained on a 15K human-annotated instructions dataset. We constructed training datasets by combining three domain-specific datasets (Dolly, GSM8K, and CodeAlpaca) with the Alpaca dataset {respectively}.

\paragraph{Baselines}
We compare existing base and State-of-the-Art methods for upgrading specialist models, including SFT, Mixed Training, Continued SFT, and DMT.
Consistent with current work~\cite{alpaca,xu2023baize,deepseekv2} on instruction tuning, we use standard SFT as the most basic comparison method.
Mixed Training~\cite{dong2023abilities} is another baseline that combines multiple domains (GSM, Dolly and CodeAlpaca) directly to train a model.
{Continued SFT continues training with the Mixed Training model but focuses on the filtered error data, incorporating our tag retrieval.}
DMT~\cite{dong2023abilities} learns specialized abilities first and then learns general abilities with a small amount of specialized data. Details are provided in the Appendix~\ref{evaluation_details},~\ref{training_details}, and~\ref{retrieval_details}.


\subsection{Main Results}\label{main_res}
As shown in Table \ref{main}, we present the results across multiple tasks and various baseline methods (see Appendix~\ref{appendix:general_performance} for full results).
Overall, our approach outperforms the baseline methods across different domains. 
The improvement is particularly significant on Llama-2 on the Alpaca-eval and GSM8K  benchmarks, with improvement exceeding 7\% and 4\%, respectively. This suggests that \method, by selectively enhancing training on data where the model underperforms, can achieve gains similar to human learning from mistakes, significantly improving its performance in that domain. We showcase these improvements through case studies, detailed in the Appendix \ref{appendix:case_study}.
Notably, \method does not significantly drop in general capability tasks, indicating stable performance in these areas. This also highlights the importance of retrieving additional general data similar to the erroneous cases. By incorporating more relevant similar general data, the model is able to alleviate the forgetting phenomenon associated with continuous training, thereby preserving its general capabilities.  In summary, our method improves performance on domain-specific tasks by targeting areas where the model underperforms, while maintaining stable performance on general tasks.

\subsection{Analysis on Bad Case Generation}
\begin{figure}[t!]
    \centering
\includegraphics[width=0.95\linewidth]{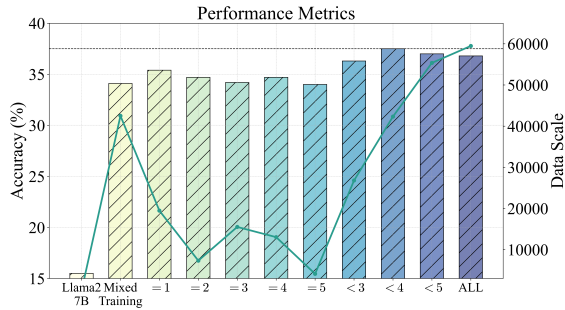}
    \caption{{The impact of different response scores and their data volume.  Selecting data with larger errors for continuous training will lead to better results and greater efficiency than merging all data (``<4'' vs. ``ALL'').}}
    \label{error_score}
\end{figure}

\paragraph{Effect of the Generation Cases}\label{generation_case}
We test the results using data at all error levels using our training objective, as shown in Figure \ref{error_score}. {``=1'' represents selecting cases where the Assessment model's scores are 1, based on the continuation of preference training after Mixed training. Similarly, ``<3'' represents selecting cases where the Assessment model's scores are less than 3, and so on. The figure's line represents the data volume used for training at each setting.}
The results indicate that data with lower scores yields better performance, suggesting that samples with noticeable or moderate errors play a crucial role in model training. 

\paragraph{Effect of the Assessment Model} To verify whether the assessment model's scoring effectively helps in filtering out low-quality responses, we replace the original assessment model with the generation model itself to score the generated samples, followed by an iterative training process. 
We conduct a comparative experiment on LLama-2-7B, as shown in Table \ref{tab:assessment_compare}.
We can observe that the model cannot evaluate the quality of its own predictions alone. Therefore, an additional assessment model is required to aid in the evaluation, much like how humans rely on a standard answer or a teacher to identify mistakes during the learning process.


\begin{table}[t]
\centering
\scalebox{0.7}{
\begin{tabular}{lcccc}
\toprule[0.5mm]
\multirow{2}{*}{ \textbf{Domain}} & \multicolumn{2}{c}{\textbf{Self}} & \multicolumn{2}{c}{\textbf{Prometheus}} \\
\cmidrule(lr){2-3} \cmidrule(lr){4-5}
 & \textbf{Domain} & \textbf{AVG} & \textbf{Domain} & \textbf{AVG} \\
\midrule
\textbf{GSM} & 34.8 & 34.6 & \textbf{39.2} & \textbf{37.2} \\
\textbf{CodeAlpaca} & 26.0 & 34.7 & \textbf{28.4} & \textbf{36.7} \\
\textbf{Dolly} & 18.7 & 34.9 & \textbf{25.0} & \textbf{36.6} \\
\bottomrule[0.5mm]
\end{tabular}
}
\caption{Comparison of assessment models in which ``Self'' refers to using the model itself as the assessment model and ``Prometheus'' refers to the assessment model employed in our method.}
\label{tab:assessment_compare}
\end{table}



\subsection{Analysis on Similar Case Retrieval}

\paragraph{Effect of Retrieval Method}
We perform experiments on LLama-2 to evaluate the retrieval step, as shown in Table \ref{retrieval}. Specifically, we compare the baseline (Mixed training) and further fine-tuning based on mixed training with error samples alone (only error) and with three retrieval strategies: mean vector, cluster-based, and our proposed tag-based similarity.
Mean Vector Similarity computes the average embedding vector~\cite{Zeng2022CurriculumLF,karpukhin-etal-2020-dense,zhao2024seer,rao2023DCD} from all error samples and uses it to perform cosine similarity~\cite{rao2022reproducibility} comparisons to retrieve relevant items. 
Cluster Based Similarity groups error samples into clusters using the K-nearest neighbors algorithm~\cite{knn,wu-etal-2023-self}, then calculates cosine similarity between the cluster centers and the candidate samples, including diverse yet similar cases.
Our results reveal that retrieving similar samples significantly improves performance compared to direct training on error samples. Among the three retrieval strategies, the performance improvement increases as the retrieval granularity becomes finer (Mean-Vector < Cluster-Based < Tag-Based )~\cite{zhao2024funnelrag}.

\paragraph{Effect of Retrieval Data Scale}
We compare the results for retrieval sizes of 1$\times$, 2$\times$, and 3$\times$ the original error data within an iteration on the final model's effectiveness, as shown in Table \ref{retrieval_size}. 
The results indicate that the model achieves optimal performance with 1$\times$ retrieval size while increasing the data volume to 2$\times$ or 3$\times$ leads to a significant decline (e.g., on Dolly and CodeAlpaca). 
This precisely indicates that for domain training, it is not the case that the more data, the better.
Overly similar or excessive samples may introduce redundancy or noise, negatively impacting the model's generalization ability.

\begin{table}[t!]
  \centering
    \scalebox{0.70}{
    \begin{tabular}{lccccc}
    \toprule
             \multicolumn{1}{l}{ \shortstack[c]{\textbf{Domain} \\ \textbf{Dataset}}} 
          & \multicolumn{1}{c}{ \shortstack[c]{\textbf{Mixed} \\ \textbf{Training}}}
          & \multicolumn{1}{c}{ \shortstack[c]{\textbf{Only} \\ \textbf{Error}}} 
          & \multicolumn{1}{c}{ \shortstack[c]{\textbf{Mean}
          \\ \textbf{Vector} }}
          & \multicolumn{1}{c}{ \shortstack[c]{\textbf{Cluster} \\ \textbf{Based} }} 
          & \multicolumn{1}{c}{ \shortstack[c]{\textbf{Tag} \\ \textbf{Based} }} \\
    \midrule
    \textbf{GSM} & 34.1 & 36.3 & 38.1 & 38.1 & \textbf{38.1}  \\
    \textbf{CodeAlpaca} & 25.0 & 26.6 & 27.7 & 26.9 & \textbf{28.2} \\
    \textbf{Dolly} & 19.1 & 23.5 & 21.4 & 21.7 & \textbf{23.7} \\
    \bottomrule
    \end{tabular}
    }
  \caption{Comparison of different retrieval methods on fine-tuning multiple domains.  {With our dis-preferred data and training method, the finer-grained retrieved samples can assist in learning about error cases.}}
  \label{retrieval}
\end{table}
\begin{table}[t!]
  \centering
    \scalebox{0.8}{
    \begin{tabular}{lccc}
    \toprule
        \textbf{Domain Dataset}
          & \textbf{1$\times$ Scale} & \textbf{2$\times$Scale} & \textbf{3$\times$Scale} \\
    \toprule
    \textbf{GSM} & \textbf{38.1} & 37.7 & 37.7 \\
    \textbf{CodeAlpaca} & \textbf{28.2} & 27.2 & 27.4 \\
    \textbf{Dolly} & \textbf{23.7} & 21.3 & 21.2 \\
    \bottomrule
    \end{tabular}
    }
  \caption{Impact of retrieval size on results. The results show that the best results are obtained by retrieving only an equal amount of data {combined with our selected
data and training method.}}
  \label{retrieval_size}
\end{table}




\begin{table}[t!]
  \centering
    \scalebox{0.7}{
    \begin{tabular}{lccccc}
    \toprule
            \textbf{Domain Dataset}
          & {\textbf{SFT}} & \textbf{DPO} & \textbf{ORPO}  & \textbf{Smaug} & \textbf{Ours} \\
    \midrule
    \textbf{GSM} & {32.6} & 36.5 & 37.2 & 33.4 & \textbf{38.1}  \\
    \textbf{CodeAlpaca} & {28.1} & 27.4 & 26.3 & 25.4 & \textbf{28.2} \\
    \textbf{Dolly} & {18.0} & 22.7 & 20.4 & 16.0 & \textbf{23.7} \\
    \bottomrule
    \end{tabular}
    }
  \caption{Comparison results of optimization goals for multiple domains. {SFT means to continue training with the error and retrieve data using SFT loss.}
  }
  \label{optimization_objective}
\end{table}
\begin{table}[t]
\centering
\scalebox{0.7}{
\begin{tabular}{lcccc}
\toprule[0.5mm]
\multirow{2}{*}{ \textbf{Domain}} & \multicolumn{2}{c}{\textbf{Pred.}} & \multicolumn{2}{c}{\textbf{\method}} \\
\cmidrule(lr){2-3} \cmidrule(lr){4-5}
 & \textbf{Domain} & \textbf{AVG} & \textbf{Domain} & \textbf{AVG} \\
\midrule
\textbf{GSM} & 35.9 & 36.1 & \textbf{39.2} & \textbf{37.2} \\
\textbf{CodeAlpaca} & 26.8 & 35.9 & \textbf{28.4} & \textbf{36.7} \\
\textbf{Dolly} & 23.5 & 36.3 & \textbf{25.0} & \textbf{36.6} \\
\bottomrule[0.5mm]
\end{tabular}
}
\caption{Exploring the impact of bad case filtering and similar case retrieval. ``Pred.'' refers to the iterative DPO process using negative generation without selection.
}
\label{tab:dpo* analysis}
\end{table}

\subsection{Analysis on Iterative Preference Training}
\paragraph{Effect of the Iteration}
\begin{figure}[h!]
    \centering
    \includegraphics[width=1\linewidth]{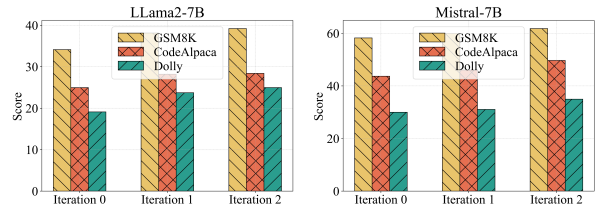}
    \caption{Domain evaluation of iteration. The iteration 0 corresponds to the  mixed training model.}
    \label{Iteration}
\end{figure}
As shown in Figure \ref{Iteration}, we show the results of the assessment of each domain after several rounds of self-iterations. 
The scores of the corresponding domains on the 3 domains can be further improved after iterative training. 
This improvement is most obvious the first time, and the improvement in the subsequent iteration process is less relative to the first iteration. 
This shows the generalization ability of ours in multiple domains and the effectiveness of the iterative operation.

\begin{figure}[t!]
    \centering
    \includegraphics[width=0.85\linewidth]{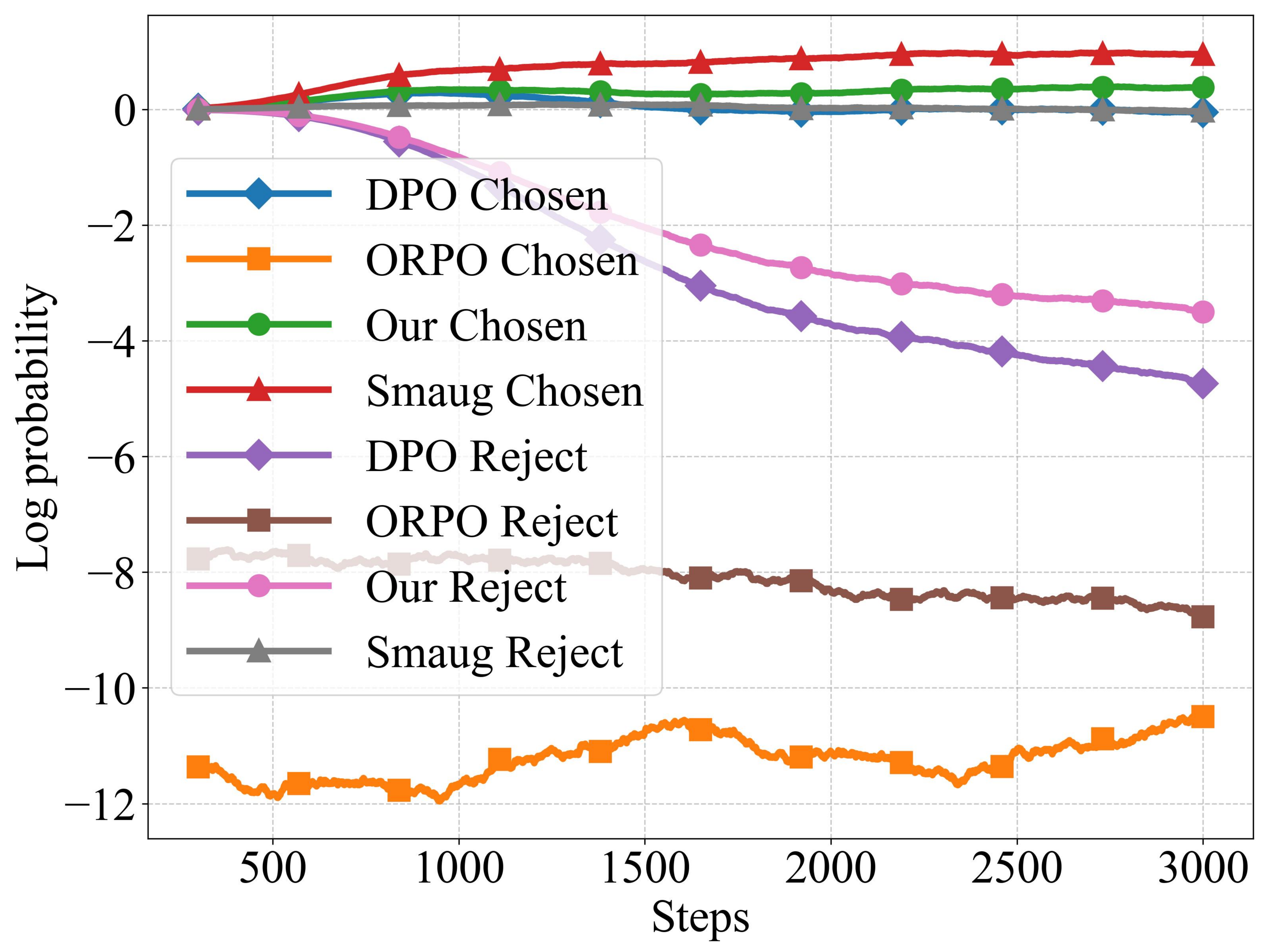}
    \caption{Comparing multiple different preference optimization objectives. The results show that our optimization objective better improves the probability of chosen samples while reducing the probability of rejected ones.}
    \label{loss}
\end{figure}

\paragraph{Effect of the Optimization Objective}
We compare here several variant approaches to DPO~\cite{dpo}, namely ORPO~\cite{hong2024orpo}, eliminating the necessity for an additional preference alignment phase and Smaug~\cite{pal2024smaug}, which adds a regular term to mitigate the reduced probability of chosen in the original DPO. 
From the training results and the average results for each domain of LLama-2 shown in Table \ref{optimization_objective}, introducing the SFT loss constraint can significantly help the model to strengthen the compliance of the instructions and thus obtain a better domain effect in three domains. 
We show that models trained with our objective learned to reflect the preference throughout the training process.
As shown in Figure \ref{loss}, we compare the training curves of the logarithmic probabilities chosen and rejected for each optimization objective.
As the training progresses, the probability of choosing increases, and the probability of rejected responses decreases, suggesting that our method successfully preserves the contribution of SFT samples to the training while reducing the occurrence of unwanted responses in the model unlike DPO is unchanged. 

\begin{figure}[t!]
    \centering
\includegraphics[width=1\linewidth]{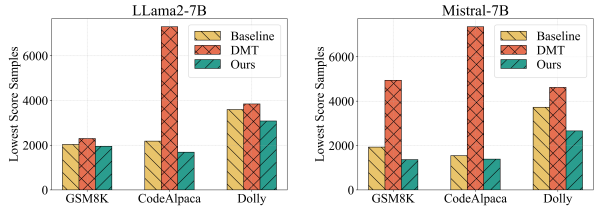}
    \caption{Comparison of the Number of Error Cases across Methods. After our method's training, the model's error samples were significantly reduced compared to the baseline and DMT.}
    \label{case number}
\end{figure}

\begin{figure}[t!]
    \centering
    \includegraphics[width=1\linewidth]{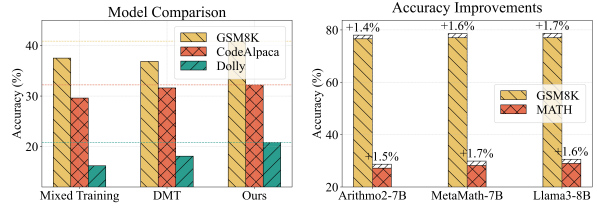}
    \caption{Scalability on Stronger Models.
    (a) Applying APT to a broader set of domain models enhances the model's capabilities. (b) Extending APT to advanced mathematical models further improves performance.
    }
    \label{scale_all}
\end{figure}

\paragraph{Effect of the Bad Case Filtering and Similar Case Retrieval}
As shown in Table \ref{tab:dpo* analysis}, merely generating negative samples to construct DPO data for iterative preference training fails to significantly enhance model performance. Instead, this training process benefits from the entire method, which includes filtering bad cases and retrieving similar cases. 
The combination of all the components finally results in the best performance.

\subsection{Method Robustness}
\paragraph{Training Robustness}

To better illustrate how \method leverages erroneous samples identified by the assessment model to enhance the model's performance,
We show the number of erroneous samples, in which assessment model judgment is less than 4 scores, before (Baseline) and after training (Ours) across three domains in Figure \ref{case number}.
We compare the results of the three methods, mixed training and DMT, and our method with only one iteration. 
The results show that our method significantly reduces the number of incorrect samples of responses in the training set in several domains while boosting the test set, indicating that our method indeed learns from the training samples. 

\paragraph{Model Scale Robustness}
We show scalability with larger model LLama2-13B and several leading models in mathematics, such as Arithmo2~\cite{jindal_2023_arithmo} and MetaMath~\cite{meta-math} and LLama3-8B~\cite{dubey2024llama}. 
For LLama2-13B, we repeated the experiment similar to the main table, and the results show that it still brings an improvement over the existing methods, 
as shown in Figure \ref{scale_all}.
Although these mathematical models are sufficiently trained on data from the mathematical domain, additional training on the erroneous data can still enhance their performance. 
Combining \method still improves the model's results, demonstrating the method's scalability potential in a wider range of models with greater scalable potential.





\section{Conclusion}
This work introduces a unified framework APT to enhance domain-specific LLM capabilities further. 
The framework significantly improves performance across multiple areas, including instruction-following, mathematical reasoning, and coding. 
We successfully identified model deficiencies and effectively optimized them by finding similar cases for targeted training, thereby improving the overall model performance. 
Our results demonstrate the potential for further optimization of LLMs on challenging data like humans, providing a promising solution for enhancing model capabilities.

\section*{{Limitations}}
Our current limitations are similar to those typically encountered in reinforcement learning studies~\cite{pal2024smaug,ivison2024unpacking}. Specifically, the performance of our model is primarily constrained by the quality of the scoring information provided. 
Since our approach relies heavily on the accuracy of the scoring model, any limitations in its ability to effectively evaluate our outputs may lead to a degradation of the overall iterative reward mechanism.
Furthermore, with respect to the retrieval step for error-related data, we acknowledge that our method is currently dependent on the performance of the InsTag model~\cite{lu2023instag}. 
While the retrieval process is focused on relevance, it is relatively coarse-grained and may not fully capture the necessary diversity.
Additionally, due to resource constraints, we have only conducted experiments on models up to the 13B scale and have not been able to extend our experiments to larger models, such as those with 70B parameters, to further explore the effectiveness of our method at scale.

\section*{Ethics Statement}
Our work follows the ACL Ethics Policy. 
Our findings are based on publicly available datasets for reproducibility purposes.
LLMs can contain potential racial and gender bias. 
The relevant code cannot be linked publicly (github) due to internal company constraints. 
If you need to do research, please contact the first author by email to get it.

\section*{Acknowledgments}
This work was supported in part by the National Natural Science Foundation of China (Grant No. 62206076), Guangdong S\&T Program (Grant No. 2024B0101050003), Guangdong Basic and Applied Basic Research Foundation (Grant No. 2024A1515011491), and Shenzhen Science and Technology Program (Grant Nos. ZDSYS20230626091203008, KJZD20231023094700001, KQTD2024072910215406). 
We would like to thank the anonymous reviewers and meta-reviewer for their insightful suggestions.



\bibliography{custom}
\bibliographystyle{acl_natbib}

\clearpage
\appendix
\section{Appendix}\label{sec:appendix}

\begin{algorithm}[t]
\caption{Our Proposed APT Method}\label{capt_al}
\begin{flushleft}
    \hspace*{\algorithmicindent} \textbf{Input:}  fine-tuned model  $f_{\theta}$, assessment
model $f_{\theta_0}$, labeled dataset $\mathcal{D}_{\mathrm{origin}} = \{(x^i, y^i)\}_{i=1}^L$, retrieval datasets pool $\mathcal{D}_{\mathrm{pool}} = \{(x^j, y^j)\}_{j=1}^U$

\hspace*{\algorithmicindent}\textbf{Output:} fine-tuned model $f_{\theta}$
\end{flushleft}
\begin{algorithmic}[1]
\Repeat
\Statex~~~~~~\textcolor{lightgray}{\# Bad Case Generation}
\State Generate dataset of bad cases $\mathcal{D}_{\mathrm{error}}$:
\Statex~~~~~~~~~~Generate the predictions $\{a^i\}_{i=1}^L$ using fine-tuned model  $f_{\theta}$ of questions $\{x^i\}_{i=1}^L$
\Statex~~~~~~~~~Generate the scores using assessment
model $f_{\theta_0}$ of labeled dataset and predictions $\{(x^i, y^i, a^i)\}_{i=1}^L$
\Statex~~~~~~~~~~Select low-score cases and generate $\mathcal{D}_{\mathrm{error}}= \{(x^i, y^i, a^i)\}_{i=1}^N$, where $x^i \sim \mathcal{D}_{\mathrm{origin}}$ 

\Statex~~~~~~\textcolor{lightgray}{\# Similar Case Retrieval}
\State Generate dataset of similar cases $\mathcal{D}_{\mathrm{retrieval}}$:

\Statex~~~~~~~~~~Find the similar and the same number of questions $\{x^j\}_{j=1}^M$ using $\mathcal{D}_{\mathrm{error}}$ as queries from $\mathcal{D}_{\mathrm{pool}}$  

\Statex~~~~~~~~~~Generate $\mathcal{D}_{\mathrm{retrieval}} = \{(x^j , y^j, a^j)\}_{i=j}^M$, where $x^j \sim \mathcal{D}_{\mathrm{pool}}$ and $a^j \sim f_{\theta}(x^j)$

\Statex~~~~~~\textcolor{lightgray}{\# Preference training}

\State Tune $f_{\theta}$ with Eq.\ref{we_loss} on $\mathcal{D}_{\mathrm{error}}$ and $\mathcal{D}_{\mathrm{retrieval}}$ to get new $f_{\theta}$


\Until{convergence or max iteration is reached}
\end{algorithmic}
\end{algorithm}

\subsection{APT Algorithm}
\label{sec:apt_algorithm}
The workflow is shown in Algorithm \ref{capt_al}. 
APT consists of two major components: data generation and model training. 
To initiate the process, we must gather comparative data for training. 
This involves obtaining both positive and negative examples, where the original answers are treated as preferred and the generated answers as disfavored, using model generation. 
Additionally, we retrieve samples that resemble the negative examples to expand the dataset. The current model generation outputs are also categorized as disfavored, allowing us to construct a new set of preference pairs that include both original and retrieved questions.
Model preference training includes our defined formulas and iterative operations.
Once the data has been constructed, iterative training can be initiated. In this process, the model is trained with newly generated preference data inputs in each iteration, producing a progressively stronger model.

\subsection{Statistics about Tag-Based Similarity}\label{statis_tag_info}
{The Tag-Based Similarity is the core retrieval approach in our framework, and the quality of the tags plays a crucial role in the accuracy of the retrieval process. The most frequently occurring tags in each dataset are as follows: for the Alpaca dataset, the most common tag is ``Information Retrieval''; for GSM8k, it is ``Mathematics''; for Code\_Alpaca, the primary tag is ``Program''; and for Dolly, it is also ``Information Retrieval''. In the Retrieval Pool, the tag ``Translation'' is most common.

\subsection{Case Study about Tag-Based Similarity}
{For each domain, we will provide two pairs of related data, with one from the retrieval pool and the other from the domain-specific data, and display their respective tag annotations.
}

\subsection{Evaluation Details}\label{evaluation_details}
\paragraph{Evaluations}
{We use MMLU~\cite{MMLU}, BBH~\cite{bbh}, ARC~\cite{clark2018think}, BoolQ~\cite{clark2019boolq}, OpenBookQA~\cite{OpenBookQA2018}, and WinoGrande~\cite{WinoGrande} to evaluate general capabilities, similar to previous works~\cite{flan-moe, wang2023far, longpre2023flan}. We evaluate the general performance of the model by computing and reporting the average scores across these datasets.}
We select three domain evaluation benchmarks (GSM~\cite{gsm}, Human-eval~\cite{humaneval} and Alpaca-eval~\cite{dubois2024alpacafarm}) to test reasoning, coding, and the following instructions. 

\paragraph{Generic Capacity.} These evaluation benchmarks test the model’s ability to handle a diverse range of question types and domains, comprehensively measuring the model's generalization ability. MMLU~\cite{MMLU} consists of a series of questions, ranging from basic to professional levels, across 57 academic subjects.
    Its multiple-choice format facilitates a relatively straightforward testing process.
    We use the official MMLU evaluation script and prompts\footnote{https://github.com/hendrycks/test}, with modifications to allow for batch processing. We evaluate using five few-shot examples, following the original setup of MMLU. We follow the setup described in the original paper of BBH~\cite{bbh}, and evaluate with chain-of-thought (CoT). The prompts were officially provided with three short in-context examples. 
     For the CoT setup, we extract the first word after the phrase ‘So the answer is,’ or the entire response if no such substring is present. {For ARC, BoolQ, OpenBookQA, and WinGrande tasks, 
we use EleutherAI LM Harness~\cite{eval-harness} for evaluation and follow its default settings for the assessment.}
Specifically, the ARC (AI2 Reasoning Challenge) dataset~\cite{clark2018think} provides both easy and challenge-level questions to assess the model's performance in reasoning tasks of varying difficulty. 
BoolQ~\cite{clark2019boolq} tests the model’s ability to answer yes/no questions based on context, while OpenBookQA evaluates its understanding of scientific knowledge. 
WinoGrande~\cite{WinoGrande} focuses on testing the model’s commonsense reasoning ability. 
    \textbf{Coding.} 
    For evaluating the coding capabilities of the models, we employ the HumanEval dataset presented in the Codex paper \cite{humaneval}. This dataset encompasses 164 programming challenges, wherein models are prompted to finalize a Python function based on its provided docstring. Following the original paper, we calculate the pass@10 to gauge the functional accuracy of the models' outputs.
     \textbf{Math Reasoning.}  The samples of GSM are divided into 7K training and 1K test problems.
    We report the results of the exact match.
     \textbf{Instruction Following.} 
    {We use AlpacaEval~\cite{alpaca_eval} to evaluate the model's instruction-following capabilities, employing GPT-3.5 as the evaluation model during the assessment.}
    
\subsection{Training Details}\label{training_details}
For the domain of the SFT model and the baseline mixed training, we utilized the LLama2-7B base and Mistral-7B with a learning rate set to $5 \times 10^{-6}$/$1\times 10^{-6}$ and a batch size of 128. For the DMT and our method, we use the LoRA~\cite{hu2022lora} fine-tuning initialization from the baseline model.
The entire training was conducted without any weight decay. We applied a linear learning rate schedule, incorporating a warmup phase where the warmup proportion was 3\% of the total training steps. 
In the preference optimization for the LLama2-7B and Mistral-7B using LoRA tuning, we adjusted the learning rate to $1 \times 10^{-6}$/$2 \times 10^{-7}$ with a reduced batch size of 32, while still omitting weight decay. The same linear learning rate schedule was employed with a 3\% warmup rate. {Based on the experimental results, as shown in Figure \ref{hyperparameters}, the regularization parameter {in Equation \ref{we_loss}} for the preference loss is set to 0.5.}

\begin{figure}[t!]
    \centering
    \includegraphics[width=0.8\linewidth]{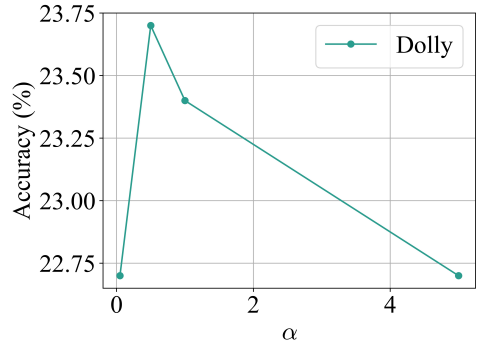}
    \caption{
The exploration of hyperparameters.}
    \label{hyperparameters}
\end{figure}

\subsection{Retrieval Details}\label{retrieval_details}
{Our retrieval method builds on the error-preference dataset generated during the bad case generation step. Using the tag-based similarity method, we select an equal amount of retrieval data from the retrieval pool to supplement the error-preference dataset, as justified in the ablation study (Figure \ref{error_score}). For example, combining GSM8K (7K) and Alpaca (52K), we identified 42K low-scoring samples (scores < 4) and retrieved an additional 42K samples from the retrieval pool, creating an 84K preference dataset for optimization in the current iteration.
}

\subsection{General Capability Experiments}\label{appendix:general_performance}
{In this work, we assess the general capabilities of the model using several widely recognized benchmark tasks: MMLU, BBH, ARC, BoolQ, OpenBookQA, and WinoGrande. 
To quantify the model’s overall capability and detect any potential forgetting, we report the average performance across these tasks, providing an aggregate measure of the model's general competence. 
The results are shown in Table \ref{General_capability}. 
The overall performance of our method demonstrates a slight improvement over the Mixed Training baseline, suggesting that additional domain-specific training does not lead to forgetting in our approach.}


\begin{table*}[t]
  \centering
  \scalebox{0.75}{
    \begin{tabular}{lllllccccc}
    \toprule
    \multicolumn{1}{l}{\textbf{Method}} & \multicolumn{1}{p{4.225em}}{\textbf{Domain}\newline{}\textbf{Dataset}} & \multicolumn{1}{p{4.225em}}{\textbf{MMLU}} & \multicolumn{1}{l}{\textbf{BBH}} & \multicolumn{1}{l}{\textbf{ARC-e}} & \multicolumn{1}{l}{\textbf{ARC-c}}& \multicolumn{1}{l}{\textbf{Boolq}} & \multicolumn{1}{l}{\textbf{OpenBookQA}} & \multicolumn{1}{l}{\textbf{WinoGrande}} & \multicolumn{1}{l}{\textbf{Avg.}} \\
    \midrule
    \multicolumn{1}{l}{\textbf{LLama2-7B}} & - & 45.2 & 40.7 & 74.6 & 46.3 & 77.7 & 44.2 & 69.3 & 56.9 \\
    \midrule
    \multirow{3}[2]{*}{\textbf{SFT}} 
        & GSM        & 46.7 & 39.2 & 76.8 & 49.7 & 79.9 & 47.4 & 69.7 & 58.5 \\
        & CodeAlpaca & 47.2 & 40.6 & 76.9 & 50.9 & 79.3 & 46.8 & 69.1 & 58.7 \\
        & Dolly      & 48.2 & 39.3 & 78.3 & 52.5 & 79.3 & 47.0 & 69.5 & 59.1 \\
    \midrule
    \textbf{Mixed Training} & - & 45.6 & 39.2 & 79.0 & 53.1 & 80.2 & 47.0 & 69.6 & 59.1 \\
    \midrule
    \multirow{3}[2]{*}{\textbf{\quad+Continued SFT}} 
        & GSM        & 45.8 & 39.4 & 78.5 & 51.9 & 80.2 & 47.2 & 69.9 & 59.0 \\
        & CodeAlpaca & 45.6 & 39.8 & 78.2 & 51.3 & 80.4 & 46.8 & 70.3 & 58.9 \\
        & Dolly      & 45.5 & 40.0 & 78.8 & 52.4 & 80.5 & 46.8 & 69.6 & 59.1 \\
    \midrule
    \multirow{3}[2]{*}{\textbf{\quad+DMT}\small{~\cite{dong2023abilities}}} 
        & GSM        & 45.4 & 37.8 & 78.7 & 53.0 & 81.4 & 47.2 & 69.8 & 59.0 \\
        & CodeAlpaca & 45.5 & 37.8 & 76.9 & 51.4 & 80.7 & 47.4 & 69.5 & 58.5 \\
        & Dolly      & 45.3 & 38.1 & 77.0 & 50.9 & 80.6 & 47.2 & 69.9 & 58.4 \\
    \midrule
    \multirow{3}[2]{*}{\textbf{\quad+Ours}} 
        & GSM        & 45.4 & 39.4 & 78.9 & 52.6 & 81.0 & 48.0 & 70.9 & \textbf{59.4} \\
        & CodeAlpaca  & 45.2 & 39.7 & 78.7 & 52.7 & 80.7 & 48.0 & 70.2 & \textbf{59.3} \\
        & Dolly      & 45.1 & 39.0 & 79.4 & 53.4 & 80.5 & 47.8 & 69.9 & \textbf{59.3} \\
    \midrule
    \textbf{Mistral-7B-V0.3} & - & 62.5 & 58.0 & 78.4 & 52.6 & 82.2 & 44.0 & 73.6 & 64.5 \\
    \midrule
    \multirow{3}[2]{*}{\textbf{SFT}} 
        & GSM        & 61.9 & 58.2 & 81.8 & 56.4 & 84.5 & 46.8 & 75.5 & \textbf{66.5} \\
        & CodeAlpaca & 62.1 & 58.2 & 80.4 & 56.1 & 84.5 & 46.4 & 75.3 & 66.1 \\
        & Dolly      & 60.7 & 58.8 & 82.1 & 58.2 & 85.1 & 47.0 & 75.5 & \textbf{66.8} \\
    \midrule
    \textbf{Mixed Training} &-& 61.0 & 58.5 & 81.7 & 57.3 & 85.6 & 46.4 & 74.4 & 66.4 \\
    \midrule
    \multirow{3}[2]{*}{\textbf{\quad+Continued SFT}} 
        & GSM        & 60.8 & 58.5 & 80.9 & 56.7 & 85.4 & 46.2 & 75.3 & 66.3 \\
        & CodeAlpaca & 61.0 & 58.5 & 81.2 & 56.7 & 85.5 & 46.2 & 74.8 & 66.3 \\
        & Dolly      & 61.0 & 58.2 & 81.3 & 56.6 & 85.4 & 46.2 & 75.1 & 66.3 \\
    \midrule
    \multirow{3}[2]{*}{\textbf{\quad+DMT}\small{~\cite{dong2023abilities}}} 
        & GSM        & 61.0 & 59.5 & 79.0 & 55.4 & 85.9 & 46.8 & 74.7 & 66.1 \\
        & CodeAlpaca & 60.9 & 58.5 & 78.6 & 55.3 & 86.2 & 47.2 & 74.7 & 65.9 \\
        & Dolly      & 61.0 & 59.6 & 78.9 & 55.6 & 86.0 & 46.6 & 74.7 & 66.0 \\
    \midrule
    \multirow{3}[2]{*}{\textbf{\quad+Ours}} 
        & GSM         & 60.8 & 58.2 & 80.6 & 57.9 & 86.5 & 46.6 & 75.0 & \textbf{66.5} \\
        & CodeAlpaca & 60.9 & 59.1 & 81.6 & 57.5 & 86.0 & 46.8 & 74.7 & \textbf{66.7} \\
        & Dolly     & 60.7 & 58.8 & 81.9 & 58.5 & 86.2 & 46.4 & 74.8 & \textbf{66.8} \\
    \bottomrule
    \end{tabular}}
    \caption{The evaluation results of comprehensive performance across multiple tasks on several LLMs show that our approach does not lead to any degradation in performance in the general domain.}
  \label{General_capability}
\end{table*}


\subsection{Prompts}\label{Judge_prompt}
The following prompt is the built-in evaluation prompt of Prometheus 2-7B, used to assess the model's responses.

\subsection{Case Study}\label{appendix:case_study}
We present detailed case studies.
GSM8K rectifies logical errors in reasoning chains to derive accurate final answers. 
On CodeAlpaca, the model fulfills task requirements more precisely, such as outputting the first ten Fibonacci numbers instead of just the tenth. 
On Dolly, it enhances its commonsense knowledge, corrects errors in commonsense reasoning, and provides more detailed and accurate answers.
The mixed training answer represents the base model's response, while our answer represents the response of the model after being trained on error cases. 
In the mixed training answer, the red text highlights the incorrect parts of the model's response, while in our answer, the red text shows the corrected responses made by the model.

\onecolumn 



\vspace{+\baselineskip} 

\begin{tcolorbox}[colback=blue!5!white, colframe=blue!75!black, title=Prompt for assessing the model's prediction]
    \#\#\#Task Description:
    An instruction (might include an Input inside it), a response to evaluate, and a score rubric representing a evaluation criteria are given.
    
    1. Write a detailed feedback that assess the quality of the response strictly based on the given score rubric, not evaluating in general.
    
    2. After writing a feedback, write a score that is an integer between 1 and 5. You should refer to the score rubric.
    
    3. The output format should look as follows: "(write a feedback for criteria) [RESULT] (an integer number between 1 and 5)"
    
    4. Please do not generate any other opening, closing, and explanations.
    \\
    \newline
    \#\#\#The instruction to evaluate:
    
    \{instruction\}
    \\
    \newline
    \#\#\#Response to evaluate:
    
    \{response\}
    \\
    \newline
    \#\#\#Reference Answer (Score 5):
    
    \{reference\_answer\}
    \\
    \newline
    \#\#\#Score Rubrics:
    You need to score the output of the model in terms of factual correctness, meeting the user's needs, logical coherence, and completeness.
    
    Score 1: Answers provide inaccurate or incorrect information and do not fulfil the purpose and need of the user to ask the question, answers are not consistent overall or there are direct logical inconsistencies in different parts of the answer, answers do not provide enough information and important aspects are omitted.
    
    Score 2: Answers provide inaccurate or incorrect information, fulfil the purpose and need of some users to ask questions, answers are consistent overall but there are logical inconsistencies between different parts, answers do not provide enough information and important aspects are omitted.
    
    Score 3: The response provided inaccurate information, met the purpose and needs of the question posed by some users, fulfilled the formatting requirements of the question, and the overall logical coherence of the question was good, but the response did not provide enough information and omitted important aspects.
    
    Score 4: The information provided in the response is accurate, fulfils the purpose and need of the question posed by some users, fulfils the formatting requirements of the question, the overall logical coherence is excellent, and the response provides sufficient information.
    
    Score 5: The information provided in the responses was accurate and based on credible facts or data, the purpose and need of the questions posed by some users and the format of the questions were fully met, the overall logical coherence was excellent, and the responses provided sufficient information and detail.
    \\
    \newline
    \#\#\#Feedback: 
\end{tcolorbox}

\begin{tcolorbox}[
    colback=blue!5!white,
    colframe=blue!75!black,
    title=Retrieval Case on gsm8k,
    boxsep=2pt,           
    before skip=0pt,      
    after skip=40pt,       
]

\textbf{Retrieval data question 1:} Evie is collecting seashells while at the beach. Each day she collects her favorite 10 shells. At the end of 6 days, she gives 2 shells to her brother. How many shells does she have left?
\textbf{Retrieval data tag 1:} ['mathematics', 'word problem', 'arithmetic', 'problem solve', 'counting', 'logic', 'mathematical operation', 'quantity', 'mathematical equation', 'mathematical reasoning']
\textbf{Domain data question 1:} Weng earns \$12 an hour for babysitting. Yesterday, she just did 50 minutes of babysitting. How much did she earn?
\textbf{Domain data tag 1:} ['mathematics', 'arithmetic', 'word problem', 'currency conversion', 'time conversion']
\textbf{Retrieval data question 2:} Task: Evaluate the statement: Humans will never colonize Mars.
\textbf{Retrieval data tag 2:} ['evaluation', 'statement analysis', 'inference']
\textbf{Domain data question 2:} In 2021, Wayne is 37 years old. His brother Peter is 3 years older than him and their sister Julia is 2 years older than Peter. What year was Julia born in?
\textbf{Domain data tag 2:} ['mathematics', 'logic', 'arithmetic', 'problem solve', 'data analysis', 'mathematical operation', 'word problem', 'age calculation', 'comparison', 'inference']

\end{tcolorbox}

\begin{tcolorbox}[
    colback=green!5!white,
    colframe=green!75!black,
    title=Retrieval Case on Dolly,
    boxsep=2pt,           
    before skip=0pt,      
    after skip=40pt,       
]
\textbf{Retrieval data question 1:} Come up with two solutions to this equation: 3x+2y=7.
\textbf{Retrieval data tag 1:} ['mathematics', 'problem solve', 'algebra', 'equation solve', 'mathematical operation', 'mathematical equation']
\textbf{Domain data question 1:} Let's imagine I create a fake currency called Yarns with two types of coins: a super-yarn (worth 10 yarns) and a mini-yarn (worth 0.5 yarn). How much money do I have in total if I have 2 super-yarns and three mini-yarns?
\textbf{Domain data tag 1:} ['mathematics', 'currency', 'quantity', 'calculation', 'output']
\textbf{Retrieval data question 2:} Write a code snippet that uses a for loop to print the numbers 1 to 10.
\textbf{Retrieval data tag 2:} ['program', 'loop', 'printing', 'number']
\textbf{Domain data question 2:} Instruction: How do compilers use IR? Input: An intermediate language is the language of an abstract machine designed to aid in the analysis of computer programs. The term comes from their use in compilers, where the source code of a program is translated into a form more suitable for code-improving transformations before being used to generate object or machine code for a target machine.
\textbf{Domain data tag 2:} ['computer science', 'program', 'translation', 'object-oriented programming']

\end{tcolorbox}

\begin{tcolorbox}[
    colback=yellow!5!white,
    colframe=yellow!75!black,
    title=Retrieval Case on CodeAlpaca,
    boxsep=2pt,           
    before skip=0pt,      
    after skip=40pt,       
]
   
\textbf{Retrieval data question 1:} Can you tell me how to check if an input is in a list in Python?
\textbf{Retrieval data tag 1:} ['python programming', 'list manipulation', 'input validation']
\textbf{Domain data question 1:} Instruction: Edit the Python code so that it prints out the total number of elements in the list. Input: list = [1, 2, 3, 4, 5]
\textbf{Domain data tag 1:} ['python programming', 'list manipulation', 'output printing']
\textbf{Retrieval data question 2:} Write a code snippet that uses a for loop to print the numbers 1 to 10.
\textbf{Retrieval data tag 2:} ['program', 'loop', 'printing', 'number']
\textbf{Domain data question 2:} Write a function to find the number of distinct states in a given matrix. matrix = [[1, 0, 0], [1, 0, 1], [1, 1, 1]]
\textbf{Domain data tag 2:} ['program', 'matrix manipulation', 'algorithm', 'data structure', 'problem solve', 'computational thinking', 'logic', 'mathematics', 'problem analysis', 'data analysis', 'algorithm design', 'program concept']

\end{tcolorbox}


\begin{tcolorbox}[
    colback=blue!5!white,
    colframe=blue!75!black,
    title=Dolly Case 1,
    boxsep=2pt,           
    before skip=0pt,      
    after skip=40pt,       
]
    \textbf{Question:} Below is an instruction that describes a task, paired with an input that provides further context. Write a response that appropriately completes the request.
    
    \#\#\# Instruction:
    What are the primary uses of horses according to the text?
    
    \#\#\# Input:
    Horses and humans interact in a wide variety of sport competitions and non-competitive recreational pursuits as well as in working activities such as police work, agriculture, entertainment, and therapy. Horses were historically used in warfare, from which a wide variety of riding and driving techniques developed, using many different styles of equipment and methods of control. Many products are derived from horses, including meat, milk, hide, hair, bone, and pharmaceuticals extracted from the urine of pregnant mares. Humans provide domesticated horses with food, water, and shelter as well as attention from specialists such as veterinarians and farriers.
    
    \#\#\# Response:
    
    \textbf{Mixed training answer:} Horses are used for sport, recreation, work, entertainment, and therapy.
    
    \textbf{Our answer:} Horses are used for a variety of activities, including sport competitions, non-competitive recreational pursuits, working activities such as \textcolor{red}{police work, agriculture}, entertainment, and therapy. Horses were also used in \textcolor{red}{warfare}, which led to the development of riding and driving techniques. Horses also provide humans with products such as \textcolor{red}{meat, milk, hide, hair, bone, and pharmaceuticals}.
    
    \textbf{Reference answer:} Horses can be used for recreational activities and therapy as well as working activities like police work or agriculture. Historically horses were also used in warfare. Horses can also provide products like meat, milk, hide, hair, bone or pharmaceuticals.

\end{tcolorbox}

\begin{tcolorbox}[colback=green!5!white, colframe=green!75!black, title=Dolly Case 2]
    \textbf{Question:} Below is an instruction that describes a task, paired with an input that provides further context. Write a response that appropriately completes the request.
    
    \#\#\# Instruction:
    Classify the following as types of birds or types of snakes: robin, cobra, rattlesnake, eagle, viper, raptor, bluejay, cottonmouth, copperhead, sparrow.
    
    \#\#\# Input:
    
    \#\#\# Response:
    
    \textbf{Mixed training answer:} Birds: robin, \textcolor{red}{cobra, rattlesnake, viper}, raptor, bluejay.
    
    Snakes: cottonmouth, copperhead, \textcolor{red}{sparrow}.
    
    \textbf{Our answer:} Birds: robin, eagle, bluejay, raptor, \textcolor{red}{sparrow}
    
    Snakes: \textcolor{red}{cobra, rattlesnake, viper}, cottonmouth, copperhead.
    
    \textbf{Reference answer:} Robin, eagle, raptor, bluejay, and sparrow are all kinds of birds. Cobra, rattlesnake, viper, cottonmouth, and copperhead are several types of snakes.

\end{tcolorbox}

\begin{tcolorbox}[
    colback=yellow!5!white,
    colframe=yellow!75!black,
    title=Dolly Case 3,
    boxsep=2pt,           
    before skip=5pt,      
    after skip=40pt,       
]
    \textbf{Question:} Below is an instruction that describes a task, paired with an input that provides further context. Write a response that appropriately completes the request.
    
    \#\#\# Instruction:
    Tell me if these drinks are alcoholic or not: Coca-Cola, Mountain Dew, Budweiser, Rum, Vodka, Tomato Juice, Orange Juice, Sprite, Tequila, Margarita
    
    \#\#\# Input:
    
    \#\#\# Response:
    
    \textbf{Mixed training answer:} \textcolor{red}{Coca-Cola, Mountain Dew are alcoholic drinks}. Budweiser, Rum, Vodka, Tequila are alcoholic drinks. Tomato Juice, Orange Juice, Sprite are not alcoholic drinks.
    
    \textbf{Our answer:} \textcolor{red}{Coca-Cola: Not Alcoholic} \\
    \textcolor{red}{Mountain Dew: Not Alcoholic} \\
    Budweiser: Alcoholic \\
    Rum: Alcoholic \\
    Vodka: Alcoholic \\
    Tomato Juice: Not Alcoholic \\
    Orange Juice: Not Alcoholic \\
    Sprite: Not Alcoholic \\
    Tequila: Alcoholic \\
    Margarita: Alcoholic
    
    \textbf{Reference answer:} Budweiser, Rum, Vodka, Tequila and Margarita are alcoholic beverages.

\end{tcolorbox}

\begin{tcolorbox}[
    colback=blue!5!white,
    colframe=blue!75!black,
    title=GSM8k Case 1,
    boxsep=2pt,           
    before skip=5pt,      
    after skip=20pt,       
]
    \textbf{Question:} Four people lost a total of $103$ kilograms of weight. The first person lost $27$ kilograms. The second person lost $7$ kilograms less than the first person. The two remaining people lost the same amount. How many kilograms did each of the last two people lose?

    \textbf{Mixed training answer:} Let \textcolor{red}{$x$} be the amount of weight the second person lost. The first person lost $27$ kilograms, so the second person lost \textcolor{red}{$(x + 7)$} kilograms. The first and second person lost a total of \textcolor{red}{$27 + x + 7 = 34$} kilograms. The remaining two people lost a total of \textcolor{red}{ $103 - 34 = 69$} kilograms. The last two people lost the same amount, so each of them lost \textcolor{red}{$(69 / 2 = 34.5)$} kilograms. The answer is \textcolor{red}{$34.5$}.
    
    \textbf{Our answer:} Let x be the amount of weight the second person lost. The first person lost $27$ kilograms, so the second person lost \textcolor{red}{$(27 - 7 = 20)$} kilograms. The first two people lost \textcolor{red}{a total of $27 + 20 = 47$} kilograms. The last two people lost a total of \textcolor{red}{$103 - 47 = 56$} kilograms. So the last two people lost \textcolor{red}{$(56 / 2 = 28)$} kilograms each. The answer is $28$.
    
    \textbf{Reference answer:} Second person = $(27 - 7 = 20)$ kg. $103 - 27 - 20 = 56$ kg. $56 / 2 = 28 $ kg.The last two people each lost $28$ kilograms of weight.The answer is $28$.

\end{tcolorbox}

\begin{tcolorbox}[
    colback=green!5!white,
    colframe=green!75!black,
    title=GSM8k Case 2,
    boxsep=2pt,           
    before skip=5pt,      
    after skip=40pt,       
]
    \textbf{Question:} Tim has 30 less apples than Martha, and Harry has half as many apples as Tim. If Martha has 68 apples, how many apples does Harry have?

    \textbf{Mixed training answer:} Let \textcolor{red}{$x$} be the number of apples Harry has. Tim has 30 less apples than Martha, so he has \textcolor{red}{$x - 30$} apples. Harry has half as many apples as Tim, \textcolor{red}{so he has $\frac{x - 30}{2}$} apples. Martha has 68 apples, \textcolor{red}{so Tim has $x - 30 = 68$} apples. So Harry has $\frac{68}{2} = 34$ apples. The answer is $34$.
    
    \textbf{Our answer:} Let $T$ be the number of apples Tim has. Martha has 68 apples. So \textcolor{red}{Tim has $68 - 30 = 38$ apples}. Harry has half as many apples as Tim, so \textcolor{red}{Harry has $38 \div 2 = 19$ apples}. The answer is $19$.
    
    \textbf{Reference answer:} Tim has $68 - 30 = 38$ apples.Harry has $38 \div 2 = 19$ apples.The answer is $19$.

\end{tcolorbox}

\begin{tcolorbox}[
    colback=yellow!5!white,
    colframe=yellow!75!black,
    title=GSM8k Case 3,
    boxsep=2pt,           
    before skip=5pt,      
    after skip=5pt,       
]
    \textbf{Question:} Marcy is a makeup artist and has agreed to do some makeup for her friend's wedding. The only makeup she has limited stock of is lip gloss so she counts how many tubes she needs. Each tube of lip gloss will hold enough lip gloss for 3 people's makeup. Marcy decides to bring 6 tubs of lip gloss, each of which holds 2 tubes of lip gloss, and this will be the exact amount she needs for everyone's makeup. How many people is Marcy painting with makeup?
    
    \textbf{Mixed training answer:} First find the total number of tubes of lip gloss Marcy brought: $6 tubs * 2 tubes/tub = 6 \times 2 = 12 tubes$. Then \textcolor{red}{divide the total number of tubes by the number of people each tube can paint: $12 \div 3 = 4$ people}. The answer is 4.
    
    \textbf{Our answer:} Marcy has 6 tubs of lip gloss, each with 2 tubes of lip gloss, so she has \textcolor{red}{$6 \times 2 = 12$ tubes of lip gloss}. Each tube of lip gloss is enough for 3 people's makeup, so \textcolor{red}{$12 \times 3 = 36$} people's makeup. The answer is 36.
    
    \textbf{Reference answer:} Marcy is bringing 6 tubs of lip gloss * 2 tubes of lip gloss per tub of lip gloss = $6 \times 2 = 12$ tubes of lip gloss.So she must be applying makeup to 12 tubes of lip gloss * 3 people per tube of lip gloss = $12 \times 3 = 36$ people.The answer is 36.

\end{tcolorbox}

\lstdefinestyle{pythonstyle}{
    language=Python,
    basicstyle=\ttfamily\footnotesize, 
    keywordstyle=\color{blue},
    commentstyle=\color{gray},
    stringstyle=\color{red},
    breaklines=true,                   
    breakatwhitespace=true,            
    frame=single,
    numbers=left,
    numberstyle=\tiny\color{gray},
    stepnumber=1,
    numbersep=5pt,
    showstringspaces=false,
    tabsize=4,
    captionpos=b
}
\begin{tcolorbox}[colback=blue!5!white, colframe=blue!75!black, title=CodeAlpaca Case 1]
\textbf{Question}:

\begin{lstlisting}[style=pythonstyle]
Create a program that prints the first 10 Fibonacci numbers.

\end{lstlisting}

\textbf{Mix training answer}:

\begin{lstlisting}[style=pythonstyle]
def fibonacci(n):
    a = 0
    b = 1
    if n < 0:
        print("Incorrect input")
    elif n == 0:
        return a
    elif n == 1:
        return b
    else:
        for i in range(2,n):
            c = a + b
            a = b
            b = c
        return b

if __name__ == '__main__':
    print(fibonacci(10))
\end{lstlisting}

\textbf{Our answer}:

\begin{lstlisting}[style=pythonstyle]
def fibonacci(n):
    a = 0
    b = 1
    if n < 0:
        print("Incorrect input")
    elif n == 0:
        return a
    elif n == 1:
        return b
    else:
        for i in range(2,n):
            c = a + b
            a = b
            b = c
        return b

for i in range(10):
    print(fibonacci(i))
\end{lstlisting}

\textbf{Reference answer}:

\begin{lstlisting}[style=pythonstyle]
"""
Print the first ten Fibonacci numbers
"""

def print_first_ten_fibonacci_numbers():
    """Print the first ten Fibonacci numbers."""

    n1, n2 = 0, 1
    num = 0
    while num < 10:
        print(n1)
        nth = n1 + n2
        n1 = n2
        n2 = nth
        num += 1

if __name__ == '__main__':
    print_first_ten_fibonacci_numbers()
\end{lstlisting}

\end{tcolorbox}

\begin{tcolorbox}[colback=blue!5!white, colframe=green!75!black, title=CodeAlpaca Case 2]
\textbf{Question}:

\begin{lstlisting}[style=pythonstyle]
Write a function to calculate the arithmetic mean of an array of numbers
Input: arr = [5, 8, 2, 4, 7, 10]
Output:
\end{lstlisting}

\textbf{Mix training answer}:

\begin{lstlisting}[style=pythonstyle]
5.5
\end{lstlisting}

\textbf{Our answer}:

\begin{lstlisting}[style=pythonstyle]
def mean(arr):
    sum = 0
    for num in arr:
        sum += num
    return sum / len(arr)
\end{lstlisting}

\textbf{Reference answer}:

\begin{lstlisting}[style=pythonstyle]
def calculate_mean(arr):
    """
    Calculate the arithmetic mean of an array of numbers.
    """
    if len(arr) == 0:
        return 0
    
    return sum(arr) / len(arr)
\end{lstlisting}

\end{tcolorbox}

\begin{tcolorbox}[colback=blue!5!white, colframe=yellow!75!black, title=CodeAlpaca Case 3]
\textbf{Question}:

\begin{lstlisting}[style=pythonstyle]
Edit the following function so that it counts the even numbers in the given array.

Input: arr = [1, 2, 3, 4]

Output:
\end{lstlisting}

\textbf{Mix training answer}:

\begin{lstlisting}[style=pythonstyle]
 2
\end{lstlisting}

\textbf{Our answer}:

\begin{lstlisting}[style=pythonstyle]
def count_even_numbers(arr):
    count = 0
    for num in arr:
        if num % 2 == 0:
            count += 1
    return count
\end{lstlisting}

\textbf{Reference answer}:

\begin{lstlisting}[style=pythonstyle]
def countEvens(arr):
    count = 0
    for num in arr:
        if num % 2 == 0:
            count += 1
    return count
\end{lstlisting}

\end{tcolorbox}

\end{document}